\documentclass[letterpaper]{article} 
\usepackage{aaai25}  
\usepackage{times}  
\usepackage{helvet}  
\usepackage{courier}  
\usepackage[hyphens]{url}  
\usepackage{graphicx} 
\usepackage{natbib}  
\usepackage{caption} 
\usepackage{algorithm}
\usepackage{algorithmic}

\usepackage{amssymb}
\usepackage{amsmath}
\usepackage{xcolor}
\usepackage{booktabs}
\usepackage{multirow}

\usepackage{newfloat}
\usepackage{listings}

\DeclareCaptionStyle{ruled}{labelfont=normalfont,labelsep=colon,strut=off} 
\floatstyle{ruled}
\newfloat{listing}{tb}{lst}{}
\floatname{listing}{Listing}
\nocopyright

\title{Standing on the Shoulders of Giants: Reprogramming Visual-Language Model for General Deepfake Detection}

\author {
	Kaiqing Lin\textsuperscript{\rm 1}\equalcontrib,
	Yuzhen Lin\textsuperscript{\rm 1}\equalcontrib,
	Weixiang Li\textsuperscript{\rm 1},
	Taiping Yao\textsuperscript{\rm 2},
	Bin Li\textsuperscript{\rm 1}\thanks{Corresponding author.}
}
\affiliations {
	\textsuperscript{\rm 1}Guangdong Provincial Key Laboratory of Intelligent Information Processing,  Shenzhen Key Laboratory of Media Security, and SZU-AFS Joint Innovation Center for AI Technology, Shenzhen University, Shenzhen 518060, China\\
	\textsuperscript{\rm 2}Tencent Youtu Lab\\
	linkaiqing2021@email.szu.edu.cn, libin@szu.edu.cn

}

\usepackage{bibentry}

\begin{document}
	
	\maketitle
	
	\begin{abstract}
		The proliferation of deepfake faces poses huge potential negative impacts on our daily lives. 
		Despite substantial advancements in deepfake detection over these years, the generalizability of existing methods against forgeries from unseen datasets or created by emerging generative models remains constrained.
		In this paper, inspired by the zero-shot advantages of Vision-Language Models (VLMs), we propose a novel approach that repurposes a well-trained VLM for general deepfake detection.
		Motivated by the model reprogramming paradigm that manipulates the model prediction via input perturbations, our method can reprogram a pre-trained VLM model (e.g., CLIP) solely based on manipulating its input without tuning the inner parameters.
		First, learnable visual perturbations are used to refine feature extraction for deepfake detection.
		Then, we exploit information of face embedding to create sample-level adaptative text prompts, improving the performance.
		Extensive experiments on several popular benchmark datasets demonstrate that (1) the cross-dataset and cross-manipulation performances of deepfake detection can be significantly and consistently improved (e.g., over 88\% AUC in cross-dataset setting from FF++ to WildDeepfake); (2) the superior performances are achieved with fewer trainable parameters, making it a promising approach for real-world applications.
	\end{abstract}
	
	\section{Introduction}
	Deepfake refers to a series of deep learning-based facial forgery techniques\cite{AdvancingHighFidelity2020li,DesigningOneUnified2022xu,SimSwapFasterHighQuality2024chen} that can swap or reenact the face of one person in a video to another.
	In recent years, deepfake videos (a.k.a, deepfakes) have gained substantial attention due to their potential by creating and spreading false information.
	Thus, detecting deepfakes has emerged as a crucial research topic to reduce such security risks.
	
	Existing methods treat deepfake detection as a binary classification problem and predominantly utilize CNNs (e.g., Xception or EfficientNet) as the backbones of classifier. 
	In addition, some works \cite{ThinkingFrequencyFace2020qian,FaceXRayMore2020li,MultiAttentionalDeepfakeDetection2021zhao,DeepfakeBench,DF40} propose to introduce auxiliary clues, including modalities (e.g., frequency) or supervision (e.g., forgery masks) information for learning subtle forgery artifacts.
	Despite these advancements, evaluating forgeries from unseen datasets and synthesized by unseen methods beyond the training data still poses a significant challenge for practical deepfake detection.
	
	\begin{figure}[t] 
		\centering
		\includegraphics[width=0.9\linewidth]{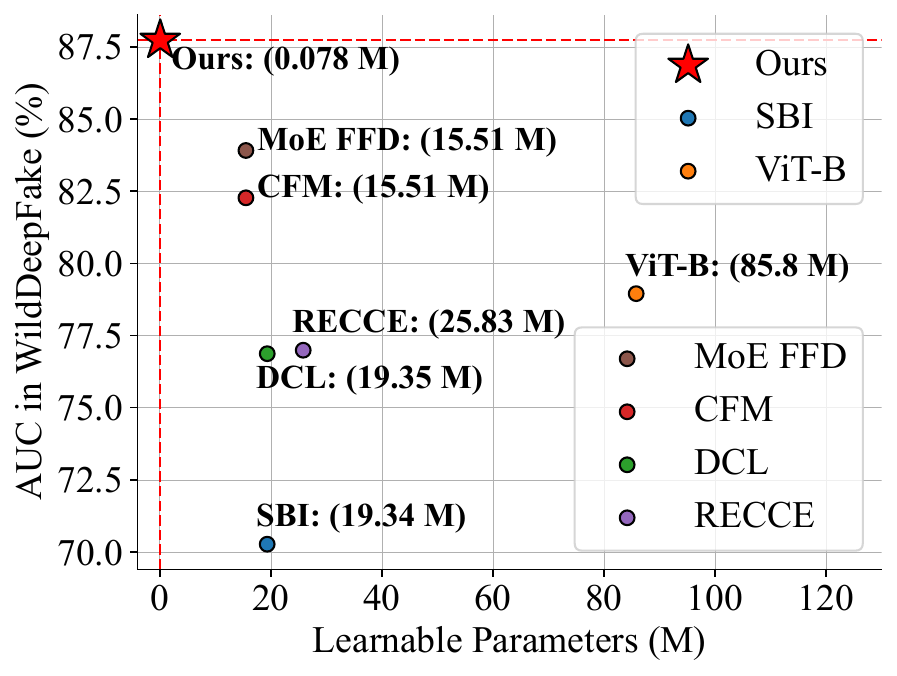}
		\caption{Comparison between our method and open-source deepfake detection models on the WildDeepfake dataset (trained on FF++). Our method with the fewest learnable parameters while achieves the best performance.}
		\label{fig:motivation}
	\end{figure}

	Vision Language Models (VLMs), such as CLIP (Contrastive Language-Image Pre-training) \cite{LearningTransferableVisual2021radford}, demonstrate robust zero-shot and few-shot generalization capabilities across diverse downstream tasks. 
	To improve deepfake detector generalization, VLFFD \cite{GeneralVisualLinguisticFace2023sun} fully fine-tunes a CLIP-based model, suggesting the effectiveness of CLIP.
	However, excessive adjustment of parameters in pre-trained models can disrupt the great pre-trained knowledge, potentially leading to suboptimal performance.
	Thus, developing a cost-efficient approach is demanding.
	
	Due to the inherent vulnerability of deep networks to adversarial attacks, a novel concept called model reprogramming \cite{AdversarialReprogrammingNeural2019elsayed,ModelReprogrammingResourceEfficient2024chen} has been proposed to re-purpose a well-trained model trained in a source domain to perform a target-domain task, just through learning universal input perturbations without modifying the source-domain model parameters.
	In other words, an additive offset applied to a deep network's input would be sufficient to adapt the network to a new task without the need of re-training or fine-tuning its inner parameters.
	The reprogramming paradigm requires significantly fewer learnable parameters compared to fine-tuning numerous parameters within the model.
	Inspired by the above advancements, we advocate that reprogramming a well-trained foundation model to identify deepfakes is a promising way to improve generalization and training efficiency.
	
	In this paper, we propose RepDFD, a novel method to reprogram a pre-trained CLIP model for effective and general deepfake detection.
	RepDFD solely learns task-specific visual perturbations (a.k.a, visual prompts) in pixel spaces for the deepfake detection task while keeping the entire CLIP model frozen. 
	Specifically, we introduce \textit{Input Transformation} to merge the image and the visual prompt and then feed it into the image encoder of CLIP. 
	Furthermore, we propose \textit{Face2Text Prompts} to generate text prompts merging information of face embedding, and then feed them into the text encoder of CLIP to guide the optimization of the visual prompts.
	This straightforward method enables the CLIP model to effectively detect deepfakes.
	Moreover, since the internal parameters are not trained, the foundation CLIP model can be reused for the other vision tasks.
	Extensive experiments on several popular deepfake benchmarks demonstrate that 
	(1) the cross-dataset and cross-manipulation performances can be significantly and consistently enhanced by equipping RepDFD for a pre-trained CLIP model; and (2) the superior performances are achieved with fewer trainable parameters (see in Fig. \ref{fig:motivation}), making it a promising approach for real-world applications.
	
	Briefly, the main contributions of this work can be summarized as follows:
	\begin{itemize}
		\item This is the first work to explore model reprogramming paradigm for deepfake detection tasks.
		\item We have proposed RepDFD to reprogram a pre-trained CLIP model by solely processing its image and text inputs without tuning the inner parameters. 
		Thus, RepDFD can be seamlessly adapted to other foundation models.
		\item We have conducted extensive experiments on several benchmark datasets, and have demonstrated that RepDFD is general and efficiency for deepfake detection.
	\end{itemize}

	\section{Related Work}
	\subsection{Deepfake Detection}
	The past five years have witnessed a wide variety of methods proposed for defending against the malicious usage of deepfakes.
	Currently, the majority of deepfake detection methods are based on deep learning, leveraging generic CNNs (e.g., Xception, EfficientNet) as the backbones of the classifier.
	Furthermore, several works \cite{ThinkingFrequencyFace2020qian,MultiAttentionalDeepfakeDetection2021zhao} utilize frequency information or localize the forged regions to improve the performance of detectors. 
	Nevertheless, the generalization challenge still hinders the application of deepfake detectors in real-world scenarios.
	To address such issue, several works \cite{FaceXRayMore2020li,DetectingDeepfakesSelfBlended2022shiohara,LAANetLocalizedArtifact2024nguyen,FakeItTill2024lin} introduce the augmented deepfake data, where two different faces are blended, during the training.
	However, all of the above methods typically involve retraining backbone networks. In general, employing more powerful backbone networks (e.g., replacing CNNs with ViTs or VLMs) produces better performances but at the cost of increased computational costs in the training stage.
	
	\subsection{CLIP Model} 
	CLIP is a vision-language model \cite{LearningTransferableVisual2021radford}, that is able to perform flexible zero-shot transfer to unseen classes using text prompts. Since CLIP is pre-trained to predict whether an image matches a textual description, it naturally fits zero-shot recognition. This is achieved by comparing image features with the classification weights synthesized by the text encoder, which takes as input textual descriptions specifying classes of interest.
	The CLIP model contains an image-encoder $E_{I}(\cdot)$ and a text-encoder $E_{T}(\cdot)$ such that the cosine similarity between the features $E_{I}(x_k)$ and $E_{T}(t_k)$ are maximized with respect to each pair $k$.
	Compared with the traditional classifier learning approach where closed-set visual concepts are learned from random vectors, vision-language pre-training allows open-set visual concepts to be explored through a high-capacity text encoder, leading to a broader semantic space and in turn making the learned representations more transferable to various vision tasks.
	Several works \cite{DEFAKEDetectionAttribution2023sha,UniversalFakeImage2023ojha,RaisingBarAIgenerated2024cozzolino} has explore that using CLIP model for detecting generated images by GANs and Diffusion Models. VLFFD (Visual-Linguistic Face Forgery Detection) \cite{GeneralVisualLinguisticFace2023sun} proposes to fully fine-tune CLIP with fine-grained text prompts for deepfake detection.
	In this work, we propose a cost-efficient method to adapt a pre-trained CLIP model for general deepfake detection.
	
	\begin{figure*}[t] %
		\centering
		\includegraphics[width=\linewidth]{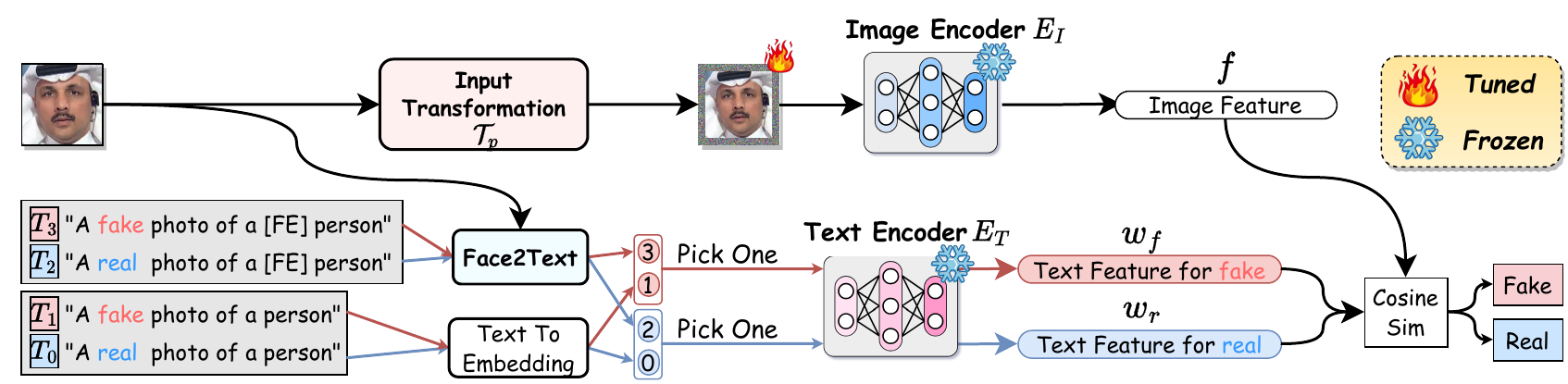}
		\caption{Overall framework of our proposed method. The core idea involves optimizing an universal visual prompt on a frozen CLIP model and generating sample-level text prompts (where the placeholder [FE] is replaced by a face embedding), aiming to adapt the model for the deepfake detection task.}
		\label{fig:pipeline}
	\end{figure*}
	
	\subsection{Model Reprogramming}
	{Parameter Efficient Fine-Tuning (PEFT) methods \cite{AutoVPAutomatedVisual2024tsao}, which significantly reduce computational and storage overheads, have garnered increasing attention.
		By fine-tuning only a limited number of additional parameters, PEFT can adapt a large pre-trained model to achieve outstanding performance on targeted tasks.
	}
	One of the PEFT methods, model reprogramming \cite{ModelReprogrammingResourceEfficient2024chen}, which incorporates prompts into inputs, offers an effective framework for repurposing models for various task-specific applications.
	The framework draws significant inspiration from adversarial reprogramming, which was first introduced by Elsayed et al \cite{AdversarialReprogrammingNeural2019elsayed}.
	Model reprogramming aims to re-use and re-align the data representation, from an existing model, for a separate task without fundamental changes to the model's inner parameters.
	This paradigm re-purposes existing knowledge by strategically transforming inputs and outputs, bypassing extensive inner model parameter fine-tuning.
	Reprogramming techniques have been widely applied in various tasks in the past few years \cite{WatermarkingReprogramming, samplespecificprompt}.
	Visual prompting (VP) \cite{ExploringVisualPrompts2022bahng} introduces a universal perturbation directly into the input data to facilitate task-specific fine-tuning while keeping the pre-trained model intact. {Although these PEFT methods have developed in other research fields, the application of PEFT methods to face forgery detection remains largely unexplored.
	In this study, we introduce a novel approach to reprogramming a pre-trained CLIP model for deepfake detection using only a few number of learnable parameters, designed to leverage the great generalization abilities of the pre-trained Vision-Language Model (VLM).
	}
	\section{Methodology}
	\subsection{Overview}
	This paper proposes RepDFD, which reprograms the pre-trained vision
	language model CLIPs for deepfake detection without altering internal parameters.
	Given a well-trained CLIP model (including the image encoder $E_I$ and the text encoder $E_T$), it introduces visual and textual prompts on the inputs for adapting the frozen $E_I$ and $E_T$ to identify deepfakes.
	To achieve this, we propose two modules, \textit{Input Transformation} and \textit{Face2Text Prompts}, which process the image and text before inputting to $E_I$ and $E_T$, respectively.
	In this way, we harness the power of the model reprograming to guide the CLIP model to focus on the deepfake detection task while preserving its inner parameters and pre-trained knowledge.
	The overall framework is depicted in Fig. \ref{fig:pipeline}. 
	In what follows, we elaborate on the details of \textit{Input Transformation}, \textit{Face2Text Prompts} and the optimization pipeline.
	
	\subsection{Input Transformation}
\begin{figure}[t]
		\centering
		\includegraphics[width=\linewidth]{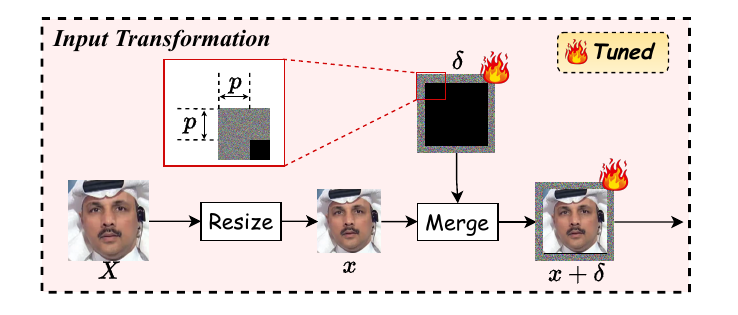}
		\caption{llustration of Input Transformation}
		\label{fig:dfvp}
	\end{figure}

	RepDFD refines the visual features from the pre-trained CLIP model for deepfake detection tasks by introducing learnable visual prompts containing perturbations that improve the visual encoder's features.
	A similar scheme like VP \cite{ExploringVisualPrompts2022bahng} and AutoVP \cite{AutoVPAutomatedVisual2024tsao} is adopted to initialize the visual prompt, placing it around images for input transformations.
	Input images are first resized smaller and subsequently incorporated with the visual prompts, aiming to match the input size of the pre-trained image encoder $E_I$.
	Fig. \ref{fig:dfvp} demonstrates the details of Input Transformation.
	Given an original image $\mathbf{X}$, the input transformation aims to merge the universal visual prompt $\boldsymbol{\delta}$ to $\mathbf{X}$.
	Formally, the process of input transformations can be described as
	\begin{equation}
		\mathcal{T}_p(\mathbf{X}, \boldsymbol{\delta}) = \text{Resize}_{p}(\mathbf{X}) + \boldsymbol{\delta},
	\end{equation}
	where $p$ is the width of the visual prompt.
	Finally, the original image $X$ of size $H\times W$ is resized to $H^{\prime} \times W^{\prime} = (H -2p)\times (W-2p)$,
	ensuring the transformed image does not overlap with $\boldsymbol{\delta}$.
	The number of trainable parameters in this method is model-agnostic, which can be computed by
	\begin{equation}\label{eq:para}
		\#Para = 3(HW - H^{\prime} W^{\prime}) = 6p(H+W) - 12p^{2}.
	\end{equation}
	This expression indicates that $\boldsymbol{\delta}$ is applied to the RGB channels of the image. Therefore, for training or inference, $\mathcal{T}_p(\mathbf{X}, \boldsymbol{\delta})$ is fed into image encoder $E_I$ to get the image feature $f$, i.e.,
	\begin{equation}
		f = E_I(\mathcal{T}_p(\mathbf{X}, \boldsymbol{\delta})).
	\end{equation}

	\subsection{Face2Text Prompts}
	To map model outputs to the target label, the CLIP model utilizes text prompts to align image features in a shared latent space, enabling classification by matching the closest text description to the image features.
	Therefore, a particular text prompt must be created for every class by injecting the information of the category.
	For instance, one of the official unified templates of text prompts is ``\textit{A photo of a [cls]}", where [cls] is a placeholder of class label.
	However, in the context of deepfake detection, the binary class labels ``real face" and ``fake face" (or synonymous terms) lack specificity compared to labels such as ``cat" or ``dog", making them difficult to accurately identify using existing text templates.
	To address this issue, VLFFD \cite{GeneralVisualLinguisticFace2023sun} introduces fine-grained text descriptions manually to fine-tune both $E_I(\cdot)$ and $E_T(\cdot)$ in the CLIP model.
	However, designing a hand-crafted text prompt that comprehensively describes forgery clues is difficult and labor-intensive.

	In this work, we aim to design a simple but effective text prompts generation method to fully harness the generalizable capabilities of the CLIP model for deepfake detection.
	The success of personalized image generation methods \cite{DreamIdentityEnhancedEditability2024chen,Arc2FaceFoundationModel2024papantoniou} has shown that introducing pre-trained image embedding as textual embeddings is beneficial for maintaining the consistency of generated subjects.
	Considering that deepfakes involve the manipulation of faces, the inconsistency of facial representations between real and fake faces is a reliable clue. 
	Inspired by this, we suggest exploiting the CLIP text encoder to integrate face embeddings into text embeddings. 
	In this way, the sample-level face information, which is difficult to express by language, can be incorporated into textual prompts, thereby enhancing the CLIP model's capability for efficient deepfake detection.

\begin{figure}[t] 
		\centering
		\includegraphics[width=\linewidth]{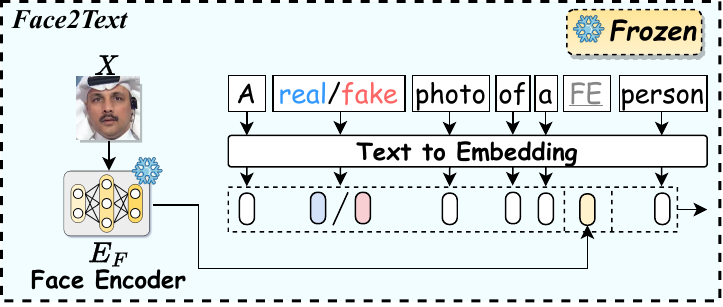}
		\caption{Illustration of Face2Text Prompts.}
		\label{fig:face-embed}
	\end{figure}

	To achieve this, a Face2Text module is proposed and the details are shown in Fig. \ref{fig:face-embed}. The prompt template $\mathbf{T}$ is designed as ``\textit{A [cls] photo of a [FE] person}".
	Given a face encoder $E_{F}$, we obtain a face embedding $S^*$ for an input image $\mathbf{X}$:
	\begin{equation}
		\centering
		\label{eq:face}
		S^* = f_{\text{map}}(E_F(\mathbf{X})).
	\end{equation}
	where $f_{map}$ represents a frozen linear layer with a random initialized, projecting the face embeddings to align the input dimension of the text encoder $E_T$. 
	By the operation $TTE$ (Text To Embedding), the text prompt $\mathbf{T}$ is initially tokenized and converted into word embeddings $\mathbf{t}$, which can be expressed as
	\begin{equation}
		\centering
		\label{eq:tte}
		\mathbf{t} = TTE(\mathbf{T}).
	\end{equation}
	Then, the placeholder token [FE] is replaced with the face embedding $S^*$, as shown below
	\begin{equation}
		\centering
		\label{eq:Face2Text}
		\mathbf{t} = Face2Text(\mathbf{t}, S^*).
	\end{equation}
	We also consider the plain prompt template ``\textit{A [cls] photo of a person}," which is not processed by Eq. (\ref{eq:Face2Text}).
	Thus, we get four different prompt templates for real and fake labels (see Tab. \ref{tab:tp}).
	An interesting finding according to the results is that an asymmetric setting for real and fake categories ($\{T_0, T_3\}$) achieves the best performance. 
	The in-depth analysis is discussed later. 
	
	Therefore, for training or inference, a text embedding $t$ is fed into text encoder $E_T$ of the pre-trained CLIP to get the text feature $w$, i.e.,
	\begin{equation}
		w = E_T(\mathbf{t}),
	\end{equation}
	
	\begin{table}
		\centering
		\begin{tabular}{cl} 
			\toprule
			Index  & Textual Template            \\ 
			\midrule
			$T_0$ & \textit{``A real photo of a person"} \\
			$T_1$ & \textit{``A fake photo of a person"} \\
			$T_2$ & \textit{``A real photo of a [FE] person"} \\
			$T_3$ & \textit{``A fake photo of a [FE] person"} \\
			\bottomrule
		\end{tabular}
		\caption{Candidate templates of text prompts.}\label{tab:tp}
	\end{table}
	
	\subsection{Optimization Pipeline}
	Following the paradigm of CLIP, the prediction probability are calculated as, 
	\begin{equation}
		\centering
		\label{eq:logit}
		P(y_i \mid x) = \frac{\exp(\cos(w_i, f) / \tau)}{\sum_{j=0}^{1} \exp(\cos(w_j, f) / \tau)},  i = \{0, 1\}
	\end{equation}
	where $cos(\cdot)$ denotes the cosine similarity, $\tau$ is the temperature parameter of CLIP, $f$ is the image feature, and $w_i$ is the text feature from the text prompt of label $y_i$.

	During training, the learnable visual prompt $\boldsymbol{\delta}$ is optimized by maximizing the likelihood of the correct label. During inference, $\boldsymbol{\delta}$ is pad around the shrunk test samples and for predictions.
	On the training dataset $\mathcal{D}$, the optimization target is to minimize $\mathcal{L}$ by tuning $\boldsymbol{\delta}$, which can be formulated as 
	\begin{equation}
		\centering
		\label{eq:ddvp}
		\mathcal{L} = \mathbb{E}_{\mathbf{X} \sim \mathcal{D}}\left[-\sum_n^N P(y_n|x_n)\log(P(y_n|x_n))\right], 
	\end{equation}
	where $N$ is the size of training dataset $\mathcal{D}$, $x_n$ is the $n$-th image sample, and $y_n$ is the ground truth label for $x_n$.
	We initialize the $\boldsymbol{\delta}$ to zero, and follow a simple gradient-based approach to directly optimize the visual prompt via back-propagation which updates $\boldsymbol{\delta}$ at step $k$ by
	\begin{equation}
		\centering
		\label{eq:update}
		\boldsymbol{\delta}^{k+1} = \boldsymbol{\delta}^{k} - \gamma {\nabla_{\boldsymbol{\delta}^k} \mathcal{L}},
	\end{equation}
	where $\gamma$ is the learning rate. 
	
	Notably, Eq. (\ref{eq:logit}) has a similar formulation to a traditional fully connected layer with input dimension $N$ and output dimension $2$ if we treat the text features $[w_0 , w_1]$ as the `weight' of a fully connected layer for the real and fake categories. Given that the adaptative text feature for each input image in our method, it implies that the model will classify samples based on sample-level dynamic classifiers. Therefore, the Face2Text module will also significantly enrich the supervised information during optimization.

	\begin{table*}
		\centering
		\small
		\begin{tabular}{lcccclcccc} 
			\toprule
			\multicolumn{5}{c}{\textbf{Frame-level}}   &  \multicolumn{5}{c}{\textbf{Video-level}} \\ 
			\cmidrule(lr){1-5} \cmidrule(lr){6-10}
			\textbf{Method} & \textbf{CDF} & \textbf{Wild} & \textbf{DFDCP} & \textbf{DFDC} & \textbf{Method} & \textbf{CDF} & \textbf{Wild} & \textbf{DFDCP} & \textbf{DFDC}  \\ 
			\midrule
			UIA-ViT (ECCV 2022)  & 82.41 & - & 75.80 & - & DCL (AAAI 2022) & 88.24 & 76.87 & 77.57 & 75.03\\
			{CFM (TIFS 2024)} & 82.78 & 78.39 & - & 75.82 & AUNet (CVPR 2023) & 92.77 & - & 86.16 & 73.82 \\
			SLADD (CVPR 2022) & 79.70 & - & - & 77.20 & SBI (CVPR 2022) & 88.61 & 70.27 & 84.80 & 71.70 \\
			FoCus (TIFS 2024) & 76.13 & 73.31 & 76.62  & 68.42 & TALL (ICCV 2023)  & 90.79 & - & - & 76.78 \\
			UCF (ICCV 2023) & 75.27 & - & 75.94 & 71.91 & TALL++ (IJCV 2024)  & 91.96 & - & - & 78.51 \\
			Ba et al. (AAAI 2024) & \textbf{86.40} & - & 85.10 & 72.10 & SeeABLE (ICCV 2023) & 87.30 & - & 86.30 & 75.90 \\
			LSDA (CVPR 2024) & 83.00 & - & 81.50 & 73.60 & LAA-Net (CVPR 2024) & 95.40 & 80.03 & 86.94 & - \\
			VLFFD (arXiv 2023) & 84.80 & 83.55 & 84.74 & - & IID (CVPR 2023) & 83.80 & - & - & 81.23 \\ 
			SA3WT (IJCV 2024) & 83.80 & -& -& 76.02 & Bi-LIG (TIFS 2024) & \textbf{97.93} & 83.00 & 91.24 & \textbf{82.57}\\   
			\midrule
			\textbf{Ours (DF)} &78.61 & \textbf{86.60} & 86.15 & 72.43 & \textbf{Ours (DF)} & 88.41 & 87.73 & 90.68 & 77.19 \\
			\textbf{Ours (FF++)} &80.00 & 85.42 & \textbf{90.57} & \textbf{77.34} & \textbf{Ours (FF++)} & 89.94 & \textbf{88.05} & \textbf{95.03} & {80.99} \\
			\bottomrule 
		\end{tabular}
		\caption{AUC (\%) of cross-datasets evaluations. The results of other SOTA methods are directly cited from their corresponding original paper. The best results are highlighted.}\label{tab:cross-data}
	\end{table*}

	\section{Experiments}
	\subsection{Experimental Settings}
	\subsubsection{Datasets}
	Following most previous works, we mainly conducted training on the FaceForensics++ (FF++) \cite{FaceForensicsLearningDetect2019rossler}. It contains 1000 Pristine (PT) videos (i.e., the real sample) and 5000 fake videos forged by five manipulation methods, i.e., Deepfakes (DF), Face2Face (F2F), FaceSwap (FS), NeuralTextures (NT) and FaceShifter (FSh). Besides, FF++ provides three quality levels in compression for these videos: raw, high-quality (HQ) and low-quality (LQ). 
	The \textbf{HQ version of FF++} is adopted by default in this paper. The samples were split into disjoint training, validation, and testing sets at the video level follows the official protocol.
	To demonstrate the performances in cross-dataset settings, four additional datasets are adopted, i.e., Celeb-DF-v2 (CDF) \cite{CelebDFLargeScaleChallenging2020li}, DeepFake Detection Challenge preview (DFDCP) \cite{DeepfakeDetectionChallenge2019dolhansky}, DeepFake Detection Challenge public (DFDC) \cite{DeepFakeDetectionChallenge2020dolhansky} and WildDeepfake (Wild) \cite{WildDeepfakeChallengingRealWorld2020zi}. 
	
	\subsubsection{Implementation details}
	We employed CLIP-ViT-L \cite{LearningTransferableVisual2021radford} as the foundation model, which is pretraiend on LAION-400M.
	A pre-trained Transface \cite{TransFaceCalibratingTransformer2023dan} was employed as $E_F$.
	The size of input for foundation model is $224\times 224$. 
	We set $p = 34$ for Input Transformations, so that trainable parameters of our method is 0.078M according to Eq. (\ref{eq:para}).
	We employed the AdamW optimizer with the learning rate 1.0, and the weight decay was fixed at 0. 
	Besides, the data preprocessing transform was as same as the original CLIP \cite{LearningTransferableVisual2021radford}
	Notably, the visual prompt $\boldsymbol{\delta}$ was initialized by zero.

	\subsubsection{Evaluation metrics}
	In this work, we mainly report the area under the ROC curve (AUC) to compare with prior works. 
	The video-level results are obtained by averaging predictions over each frame on an evaluated video. 
	To facilitate a comprehensive comparison of our method with others, we also present the results of the equal error rate (EER) in our appendix \cite{ours}.
	
	\begin{table}
		\centering
		\small
		\begin{tabular}{ccccc} 
			\toprule
			Training Data &Method  & DF & FS & FSh          \\ 
			\midrule
			\multirow{3}{*}{DF} &DCL & 99.98 & 61.01 & 68.45          \\
			&IID & 99.51 & 63.83  & 73.49          \\
			&\textbf{Ours} & 99.36 &\textbf{94.94}  & \textbf{81.51} \\
			\midrule
			\multirow{3}{*}{FS}  &DCL & 74.80 & 99.90  & 64.86  \\
			&IID & 75.39 & 99.73  & 66.18  \\
			&\textbf{Ours}    & \textbf{98.31} &99.59  & \textbf{85.21} \\
			\midrule
			\multirow{3}{*}{FSh} &DCL & 63.98        & 58.43            & 99.49         \\
			&IID & 65.42          & 59.50            & 99.50          \\
			&\textbf{Ours}    & \textbf{89.99} & \textbf{81.22} & 99.82 \\
			\bottomrule
		\end{tabular}
		\caption{AUC (\%) on cross-manipulation evaluations. The best cross-manipulation results are highlighted.}\label{tab:cross-mani}
	\end{table}

	\subsection{Comparisons with State-of-the-Arts}
	To comprehensively evaluate the generalizability of our method, we compare the performances of cross-datasets and cross-manipulation evaluations with several SOTA methods published in the past three years. 
	\subsubsection{Cross-Dataset Evaluations}
	The cross-datasets evaluation is still a challenging task because the unknown domain gap between the training and testing datasets can be caused by different source data, forgery methods, and/or post-processing. 
	In this part, we evaluate the generalization performances in a cross-dataset setting, in which detection models were trained on the FF++ (only containing DF, F2F, FS, and NT subsets for fair comparisons) and tested on other datasets.
	Our method is compared with several state-of-the-art (SOTA) methods proposed in the past three years, including: UiA-ViT \cite{UiA-ViT}, DCL \cite{DualContrastiveLearning2022sun}, CFM \cite{CFM}, AUNet \cite{Aunet}, SLADD \cite{SLADD}, SBI \cite{DetectingDeepfakesSelfBlended2022shiohara}, FoCus \cite{FoCus}, UCF \cite{UCF}, TALL++ \cite{TALL++}, TALL \cite{TALL}, \cite{ba2024exposing}, SeeABLE \cite{SeeABLESoftDiscrepancies2023larue}, LSDA \cite{LSDA}, LAA-Net \cite{LAANetLocalizedArtifact2024nguyen}, VLFFD \cite{GeneralVisualLinguisticFace2023sun}, IID \cite{IID}, SA3WT \cite{SA3WT}, and Bi-LIG \cite{Bi-LIG}.
	The experimental results in terms of frame-level and video-level AUC are shown in Tab. \ref{tab:cross-data}.
	Aside from its moderate performance on the CDF dataset, our method outperforms most competitors on the Wild, DFDC, and DFDCP datasets.
	We also report the results trained on FF++/DF, and found our method still performs better than all the competitors on the Wild, DFDC, and DFDCP datasets.
	It is important to highlight that all competitors re-train backbone networks (e.g., ResNet18 and ViT), which consist of at least 10M trainable parameters. In contrast, our method utilizes a mere 0.078M parameters, illustrating its superior general performance with fewer learnable parameters.

	\subsubsection{Cross-Manipulation Evaluations}
	Existing face forgery detectors often struggle to handle emerging manipulation techniques.
	In this part, we conduct cross-manipulation experiments involving three forgery techniques, i.e., Deepfakes (DF), FaceSwap (FS), and FaceShifter (FSh).
	We examine models trained on one manipulation type and tested across the other three.
	As shown in Tab. \ref{tab:cross-mani}, it can be observed that our method can improve cross-manipulation performances.
	These results highlight the effectiveness of our method in combating emerging unseen forgery methods.
	
	\subsection{Ablation Studies}
	In this part, we perform several evaluations to explore the effectiveness of ReDFD.
	The main results of these experiments are cross-dataset performances trained on FF++/DF.
	
	\subsubsection{Impact of reprogramming paradigm}
	In this part, we evaluate the effectiveness of our reprogramming paradigm compared with other tuning paradigms.
	Specifically, for the fine-tuning of the image feature encoder $E_I$, two established methods were assessed: Full Fine-Tuning (FFT), entailing the adjustment of all parameters within $E_I$, and Linear Probing (LP), which incorporated a learnable linear layer while maintaining $E_I$ as frozen.
	In addition, we reference a very recent work MoE-FFD \cite{MoEFFDMixtureExperts2024kong}, which presents a deepfake detection method jointly utilizing the LoRA and Adapter paradigms {to tune a frozen $E_I$}. 
	
	Beyond these three image feature extractor-related tuning paradigms, our study also incorporated a text feature extractor-related tuning paradigm, CoOp \cite{LearningPromptVisionLanguage2022zhou}, into the comparison. 
	It maintains the CLIP frozen while introducing learnable text prompts to adapt CLIP for target tasks.
	As shown in Tab. \ref{tab:tuning}, our approach outperforms other methods in most scenarios. 
	LP and CoOp inadequately facilitate the transfer of the base CLIP model to the deepfake detection task, resulting in suboptimal generalization performance. 
	We consider that these two paradigms share a common issue: they affect the utilization of image features rather than their extraction.
	Compared with them, FFT, MoE-FFD, and ours can directly impact image feature extraction, consequently improving general performance across multiple datasets.
	Furthermore, our method achieves superior generalization using significantly fewer parameters. 
	Although the number of trainable parameters does not increase, the performance gains when using a larger CLIP model with our method. 
	Conversely, the performance of MoE-FFD significantly declines on a larger model due to more trainable parameters, suggesting possible overfitting.
	We speculate that utilizing a limited number of parameters to fit deepfake-related knowledge potentially preserves the generalization capabilities of the base model.
	It reveals the potential of our approach to be effectively scaled to larger models.
	
	\begin{table}
		\centering
		\small
		\setlength{\tabcolsep}{1.7mm}
		\begin{tabular}{ccccccc} 
			\toprule
			Model & Method  &\# Para & CDF & Wild & DFDCP & Avg \\ 
			\midrule
			\multirow{4}{*}{ViT-L}
			& FFT &303M & 83.12  & 70.20 & 90.58 & 81.30 \\
			& LP &0.002M & 75.78 & 74.33 & 76.73 & 75.62 \\
			& MoE    &41.34M & 86.21 & 80.00 & 77.51 & 81.24  \\  
			& CoOp  & 0.057M & 74.72 & 74.07 & 75.82 & 74.87 \\
			& \textbf{Ours} &0.078M & \textbf{89.94} & \textbf{88.05} & \textbf{95.03} & \textbf{91.01} \\
			\midrule
			\multirow{4}{*}{ViT-B}
			& FFT &86M & 79.64 & 66.84 & 89.86 & 78.78 \\
			& LP &0.001M & 61.96 & 68.81 & 76.91 & 69.23 \\
			& MoE    &15.51M & \textbf{91.28} & \textbf{83.91} & 84.97 & 86.72 \\
			& CoOp  & 0.038M & 67.43 & 64.47 & 76.05 & 69.32 \\
			& \textbf{Ours}    &0.078M & 86.81 & 81.53 & \textbf{91.93} & \textbf{86.76} \\
			\bottomrule& 
		\end{tabular}
		\caption{
			{Comparison of AUC ($\%$) across different tuning paradigms in a cross-dataset setting. Results for MoE correspond to MoE-FFD and are sourced from the original publication. The 'Avg' column denotes the mean AUC computed over various datasets}}
		\label{tab:tuning}
	\end{table}
	
	\subsubsection{Impact of different text prompts}
	In this part, we investigate the effects of various text prompt configurations on RepDFD, including fixed text prompts, randomly initialized text prompts, and our adaptative face-related text prompts (termed \textit{dynamic text prompts}).
		It can be concluded from the experiment that the dynamic text prompts are more effective than the fixed those.
	We consider all groups of the real/fake text templates in Tab. \ref{tab:tp}, i.e., $\{T_0, T_1\}, \{T_2, T_1\}, \{T_2, T_3\}$.
	Besides, we consider a special text prompt setting, named `Rand Text', which contains two completely random initialized and frozen text embedding as long as the `Fixed Text' ($\{T_0, T_1\}$).
	As shown in Tab. \ref{tab:tp-res}, the best performance occurred when the face embeddings were only utilized in the text prompt for the fake class, corresponding to the dynamic text prompts setting $\{T_0, T_3\}$.
	{We observed that the `Fixed Text' ($\{T_0, T_1\}$) demonstrated limited effectiveness, performing similarly to `Rand Text', which lacks substantive semantic content.}
	This observation implies that semantic content in language may not significantly influence deepfake detection tasks, as the primary focus is on trace analysis.
	In contrast, integrating face embeddings into text prompts to create dynamic text prompts, corresponding to $\{T_2, T_3\}, \{T_2, T_1\}, \{T_0, T_3\}$, boosted model performance.
	It may introduce fine-grained and face-related visual information into text prompts, thereby supplementing details that are challenging to describe linguistically.
	Thus, the dynamic text prompts can not only enrich the supervision information during training, but also provide sample-level adaptative information to support classification.

	\begin{table}
		\centering
		\begin{tabular}{cccc} 
			\toprule
			Method  & CDF            & Wild           & DFDCP          \\ 
			\midrule
			Rand Text    & 82.46 & 82.30 & 89.00 \\
			$\{T_0, T_1\}$ & 82.88        & 84.91             & 88.62          \\
			$\{T_2, T_3\}$   & 85.98 & 80.85 & 87.28 \\
			$\{T_2, T_1\}$   & 85.26 & 84.80 & 87.38 \\
			\midrule
			$\{T_0, T_3\}$ (Ours)    & \textbf{88.41}         & \textbf{87.73}            & \textbf{90.68}       \\
			\bottomrule	
		\end{tabular}
		\caption{{Comparisons of AUC ($\%$) across different text prompt configurations in a cross-dataset setting. These models were trained on FF++ (DF)}}
		\label{tab:tp-res}
	\end{table}

	\begin{figure}[t] 
		\centering
		\includegraphics[width=\linewidth]{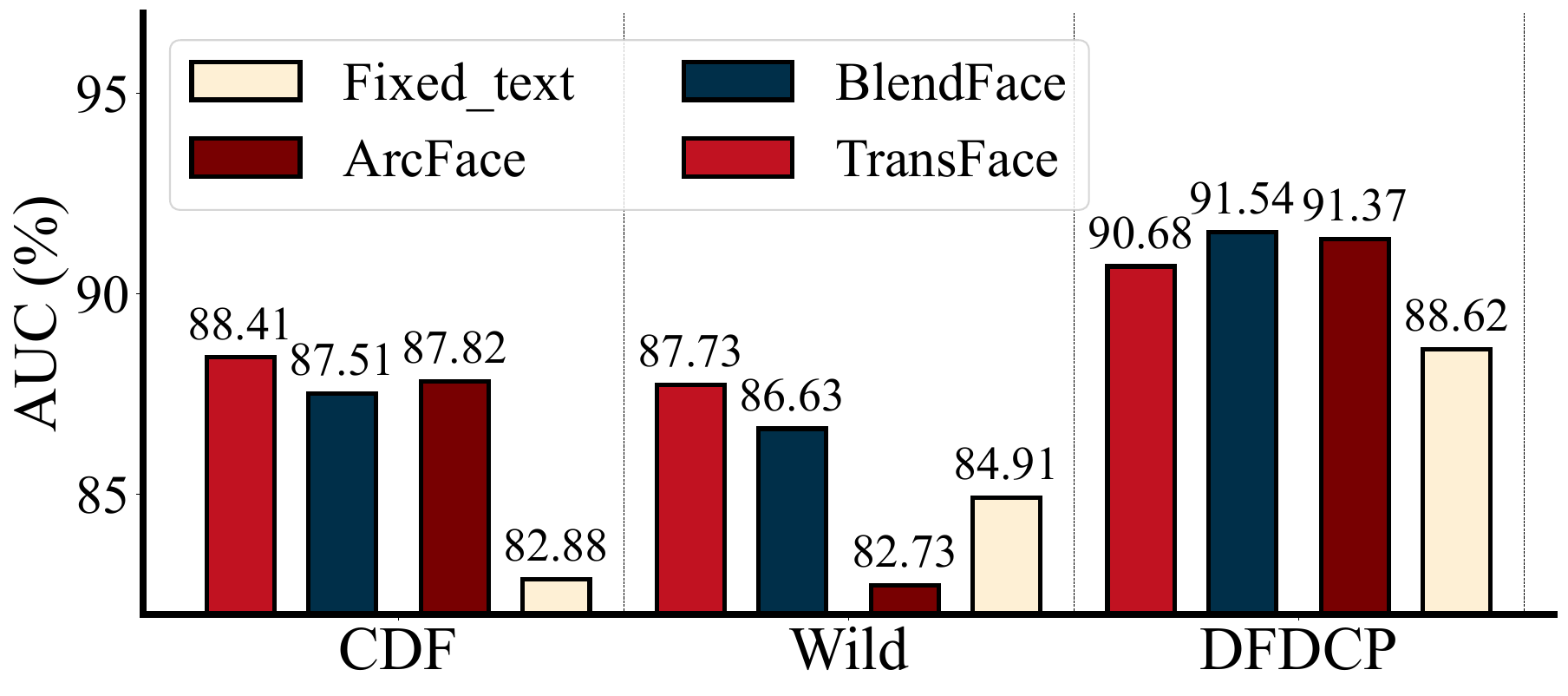}
		\caption{Comparisions of AUC ($\%$) of our method incorporating with various face embeddings. These models were trained on FF++ (DF)}
		\label{fig:face-id}
	\end{figure}
	
	\subsubsection{Impact of different face embeddings}
	To further verify the universality of Face2Text, we investigate the impact of different face embeddings on Face2Text prompts.
	We compare the results obtained by ArcFace \cite{ArcFaceAdditiveAngular2019deng}, BlendFace \cite{Blendface} and Transface \cite{TransFaceCalibratingTransformer2023dan} and fixed text (i.e., using the text prompt group $\{T_0, T_1\}$).
	As shown in Fig. \ref{fig:face-id}, our method demonstrates good performance across various face encoders $E_F$.

	\begin{table}
		\centering
		\begin{tabular}{lcccc} 
			\toprule
			Data &    $T_0$         & $T_1$           & $T_2$          & $T_3$            \\ 
			\midrule
			FF++ &  0.1889      & 0.1989              & 0.1757         &  0.2053           \\
			CDF & 0.1996        & 0.2002          & 0.1851       &  0.2021       \\
			\bottomrule	
		\end{tabular}
		\caption{The cosine similarity calculated between image features and various configurations of text prompts.}
		\label{tab:text_dis}
	\end{table}

	\subsection{Discussion}
	In this subsection, we present several in-depth analyses below, aiming to explore the effectiveness of our method.
	\subsubsection{Why are asymmetric text prompts effective?}\label{subsec:embedding}
	{In Tab. \ref{tab:tp-res}, it is noteworthy that optimal performance is achieved when the $\{T_0, T_3\}$ configuration is selected, which solely incorporates face embedding into the text for the fake label. }
	To investigate the reason, as shown in Tab. \ref{tab:text_dis}, we calculated the cosine similarity between the image features extracted by the initialized visual prompt $\boldsymbol{\delta}$ and the text prompts, both with and without face embedding.
	{Our findings indicate that cosine similarities tend to be higher for the real label ($T_0$) when using the text prompt without face embeddings, and for the fake label ($T_3$) when using the text prompt with face embeddings.}
	We speculate that text prompts with higher cosine similarities offer a better initialization for the CLIP model, enabling more effective fine-tuning of target models.
	{This observation may benefit future methodological design.}
	
	\subsubsection{The visualization of visual feature distribution}
\begin{figure}
		\centering
		\includegraphics[width=0.9\linewidth]{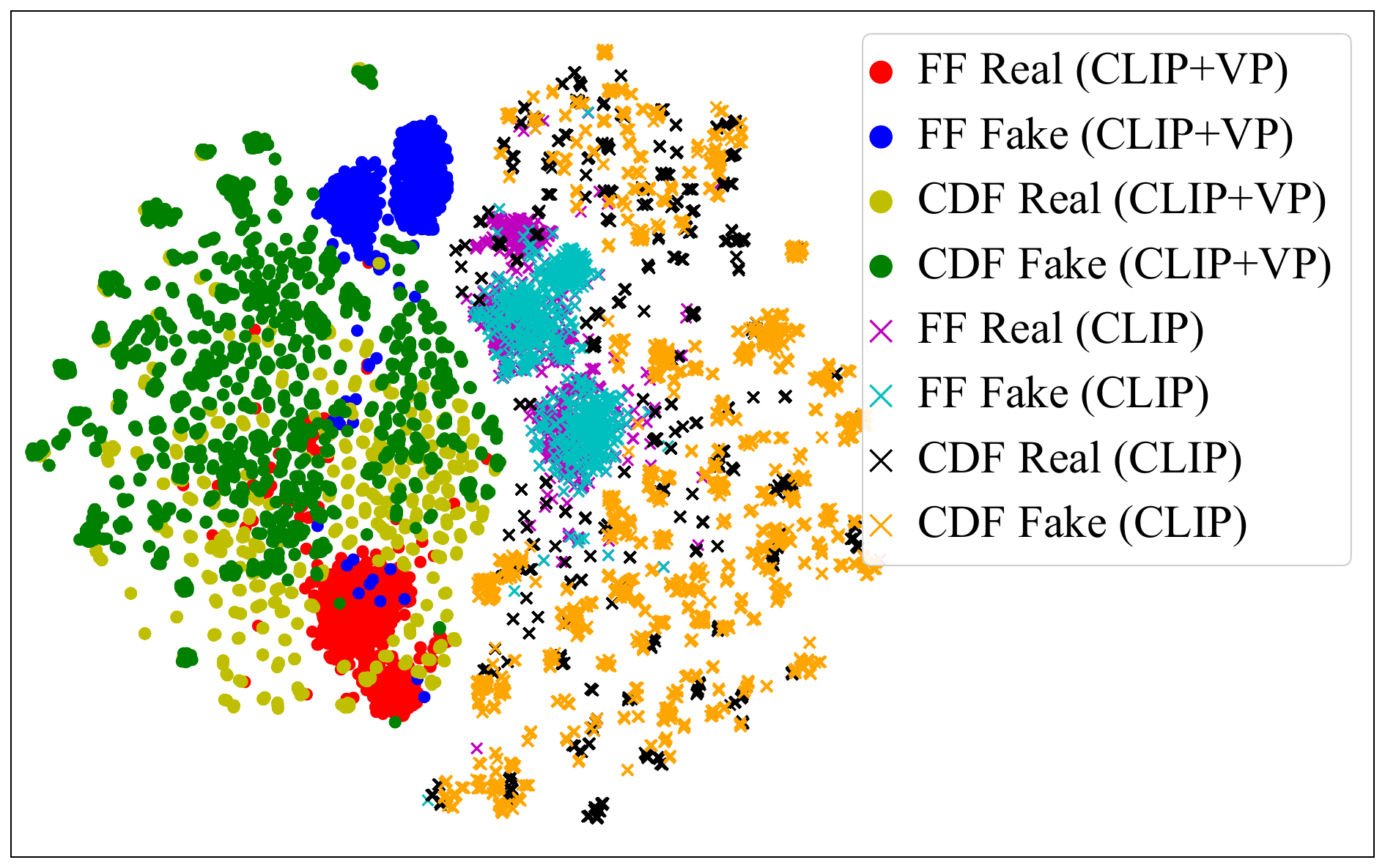}
		\caption{{The t-SNE visualization of image features. `CLIP' denotes features extracted by the original CLIP, while `CLIP+VP' refers to features extracted by the visual prompt.}}
		\label{fig:tsne}
	\end{figure}

	In this experiment, we provide the t-SNE visualization of visual feature distributions as shown in Fig. \ref{fig:tsne}. 
	The influence of the visual prompt $\boldsymbol{\delta}$ is limited exclusively to the input images and not the models, ensuring that the image features can occupy a consistent feature latent space.
	Initial observations suggest that there are significant differences in the original feature distributions between the CDF and FF datasets, which indicates the intrinsic domain discrimination capability of the unmodified CLIP. 
	Moreover, without visual prompts, the original CLIP model failed to identify real and fake images. 
	In contrast, within the FF dataset, real and fake images can be efficiently discriminated by equipping visual prompts, suggesting the effectiveness of our method.
	Furthermore, although the CDF dataset was not used during training, visual prompts trained on the FF dataset effectively endowed the model with the capability to detect deepfakes in the CDF dataset.
	We speculate that the visual prompt $\boldsymbol{\delta}$, consisting of a limited number of adjustable parameters, potentially tends to exploit the inherent capabilities of frozen models instead of introducing additional deepfake information directly, which may protect the generalization ability of pre-trained models and improve models' performance.
	Except for the observation of visual features, we also provide the visualization of text features and face embeddings in our appendix \cite{ours}, which reveals a mid-domain between different datasets in common.
	These observations are necessary to be further explored in future works.

	\section{Conclusion}
	In this paper, we have proposed RepDFD, a general but parameter-efficient method for detecting face forgeries by reprogramming a well-trained CLIP model. 
	Specifically, we employ the Input Transformation to merge the image with learnable perturbations before feeding it into the CLIP image encoder. 
	Moreover, we introduce the Face2Text Prompts to asymmetrically incorporate facial embedding information into the text prompts for real and fake categories, which are then fed into the CLIP text encoder to guide the optimization of perturbations. 
	RepDFD has effectively employed a CLIP to detect deepfakes by processing only the input images and texts, excluding the internal model parameters.
	Comprehensive experiments have demonstrated that our superior performance can be achieved with fewer trainable parameters.

	\bibliography{ref}
	\clearpage

\section*{Appendix}

We elaborate on more details and results of our work in this supplementary material.

\section{More Details of Settings}
\label{suppsec:data}
\subsection{More Implementation Details}
In the pre-processing stage, for every video frame in datasets, we employed MTCNN\cite{MTCNN} to detect and crop the facial regions, enlarged by a factor of 1.3, and subsequently resized them to 224$\times$224.

All of the code and pre-trained models of CLIP are stemmed from the official repository OpenCLIP\cite{openclip}.
For extracting face embeddings, we employed Transface\cite{TransFaceCalibratingTransformer2023dan} as the default face encoder in our method, using ViT-L version pre-trained on Glint360K\cite{an2022killing}.
It should be noticed that the dimension of face embeddings is $512$.
For CLIP-ViT-B, the face embedding can directly be integrated into text embeddings, due to the naturally alignment of dimension.
For CLIP-ViT-L, the text embedding dimension is $768$, which presents a feature integration mismatch issue.
To address the issue, a random projection layer was implemented to project the face embeddings into the target dimension. The random projection layer was initialized using the function \textbf{\textit{torch.nn.init.normal(mean=0, std=1/768)}}.

During training, the training batch size was set to $32$ and our method did not utilize any data augmentations. 
Notably, to enable Mixed Precision Training, our models were trained based on the Python library \textbf{\textit{torch.cuda.amp}}.

\subsection{More Details of Datasets}
We conduct evaluations on widely-used datasets and follow previous settings used in their corresponding datasets and compare with other methods respectively. More details on these datasets are described below.
\begin{itemize}
	\item \textbf{CelebDF (CDF)} \cite{CelebDFLargeScaleChallenging2020li} contains 590 real videos of 59 celebrities and corresponding 5639 high-quality fake videos generated by an improved forgery method. We use the stand test set consisting of 518 videos for our experiments.
	\item \textbf{DeepFake Detection Challenge Preview (DFDCP)} \cite{DeepfakeDetectionChallenge2019dolhansky} is generated by two kinds of synthesis methods on 1131 original videos. We use all 5250 videos for our experiments.
	\item \textbf{DeepFake Detection Challenge (DFDC)} \cite{DeepFakeDetectionChallenge2020dolhansky} is widely acknowledged as the most challenging dataset due to containing many manipulation methods and perturbation noises. We use the public test set consisting of 5000 videos for our experiments.
	\item \textbf{WildDeepfake (Wild)} \cite{WildDeepfakeChallengingRealWorld2020zi} contains 3805 real face sequences and 3509 fake face sequences collected from Internet. Thus, it has a variety of synthesis methods and backgrounds, as well as character identities. We use the stand test set consisting of 806 sequences for our experiments.
\end{itemize}

\section{More Experiments}
\subsection{Cross-Dataset Evaluations}
To comprehensively show the performance comparisions, we further supplement the results with the equal error rate (EER) metrics, which represents the point on the Receiver Operating Characteristic (ROC) curve where the false positive rate equals the false negative rate, providing a balanced measure of classification performance. 
As shown in Tab. \ref{tab:cross-data-eer}, the results also show promissing performance like the results with AUC metrics.
Our method's EER performance shows a consistent improvement compared to the AUC metric, highlighting the effectiveness of our approach.

\subsection{Imapct of the size of the visual prompt}
In our method, we incorporate visual prompts with input images processed through Input Transformation. Thus, the border width $p$ of visual prompts impacts both performance and the number of learnable parameters.
Herein, we conducted an ablation study to explore the imapct of $p$ was varied among the values $\{12, 23, 34, 45, 56, 67, 78\}$. 
This variation corresponds to resizing the input images to $\{90\%, 80\%, 70\%, 60\%, 50\%, 40\%, 30\%\}$ of their original size, and then pad it to original size by merging the visual prompt.
Tab. \ref{tab:width} exhibits the impact of varying $p$.
Intuitively, the learnable parameters of the visual prompt increase with an increase in $p$.
However, it is worth noting that as $p$ increases, the average generalization performances of the model initially improves but then drops significantly.
We speculate that such decline can be attributed to information loss caused by excessive scaling down of the images, suggesting a necessary trade-off between learnable parameters and the size of the input images.
Therefore, we set $p=34$ in our experiments, corresponding to a resized image that is $70\%$ of its original size.

\begin{table*}
	\centering
	\small
	\begin{tabular}{lcccclcccc} 
		\toprule
		\multicolumn{5}{c}{\textbf{Frame-level}}   &  \multicolumn{5}{c}{\textbf{Video-level}} \\ 
		\cmidrule(lr){1-5} \cmidrule(lr){6-10}
		\textbf{Method} & \textbf{CDF} & \textbf{Wild} & \textbf{DFDCP} & \textbf{DFDC} & \textbf{Method} & \textbf{CDF} & \textbf{Wild} & \textbf{DFDCP} & \textbf{DFDC}  \\ 
		\midrule
		UIA-ViT (ECCV 2022)  & - & - & - & - & DCL (AAAI 2022) & 19.12 & 31.44 & 29.55 & 30.94 \\
		{CFM (TIFS 2024)} & 24.74 & 30.79 & - & 31.67 & AUNet (CVPR 2023) & - & - & - & - \\
		SLADD (CVPR 2022) & - & - & - & - & SBI (CVPR 2022) & 19.41 & 37.63 & 25.00 & 35.27 \\
		FoCus (TIFS 2024) & - & - & -  & -  & TALL (ICCV 2023)  & - & -  & -  & - \\
		UCF (ICCV 2023) & -  & - & - & - & TALL++ (IJCV 2024)  & - & -  & - & - \\
		Ba et al. (AAAI 2024) & - & - & - & - & SeeABLE (ICCV 2023) & - & - & - & - \\
		LSDA (CVPR 2024) & - & - & - & - & LAA-Net (CVPR 2024) & - & - & - & - \\
		VLFFD (arXiv 2023) & \textbf{22.73}  & 24.40  & 23.43  & - & IID (CVPR 2023) & -  & -  & -  & -\\ 
		SA3WT (IJCV 2024) & -  & -  & -  & - & Bi-LIG (TIFS 2024) & \textbf{7.30}  & -  & 17.03  & \textbf{25.07} \\   
		\midrule
		\textbf{Ours (DF)} & 28.99 & \textbf{21.04}  & 23.29 & 34.27 & \textbf{Ours (DF)} & 20.22 & \textbf{20.45} & 17.77 & 30.00 \\
		\textbf{Ours (FF++)} & 27.44 & 21.60 & \textbf{18.03}  & 30.32 & \textbf{Ours (FF++)} & 17.98  &  20.71 & \textbf{13.00} & 28.32 \\
		\bottomrule 
	\end{tabular}
	\caption{EER (\%) of cross-datasets evaluations. The results of other SOTA methods are directly cited from their corresponding original paper. The best results are highlighted.}\label{tab:cross-data-eer}
\end{table*}

\subsection{The visualization of textual feature distribution}
\begin{figure}
	\centering
	\includegraphics[width=\linewidth]{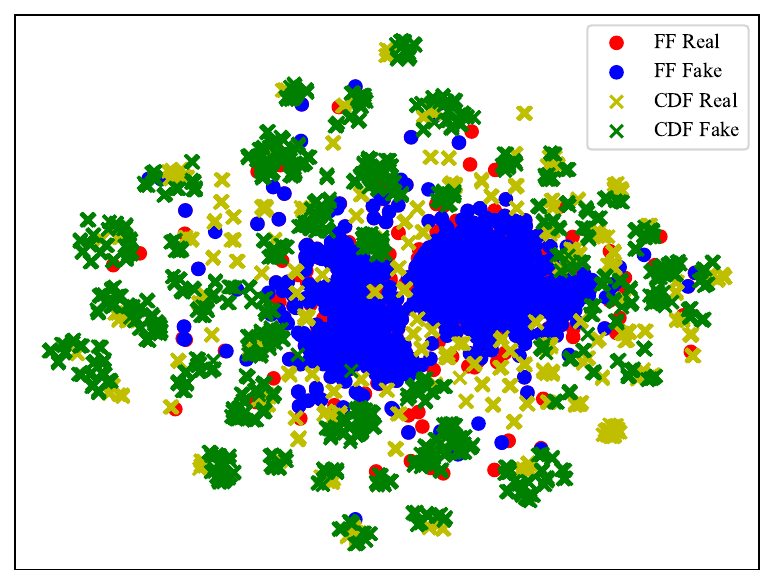}
	\caption{{The t-SNE visualization of facial features in FF and CDF datasets.}}
	\label{fig:tsne_face}
\end{figure}

\begin{figure}
	\centering
	\includegraphics[width=\linewidth]{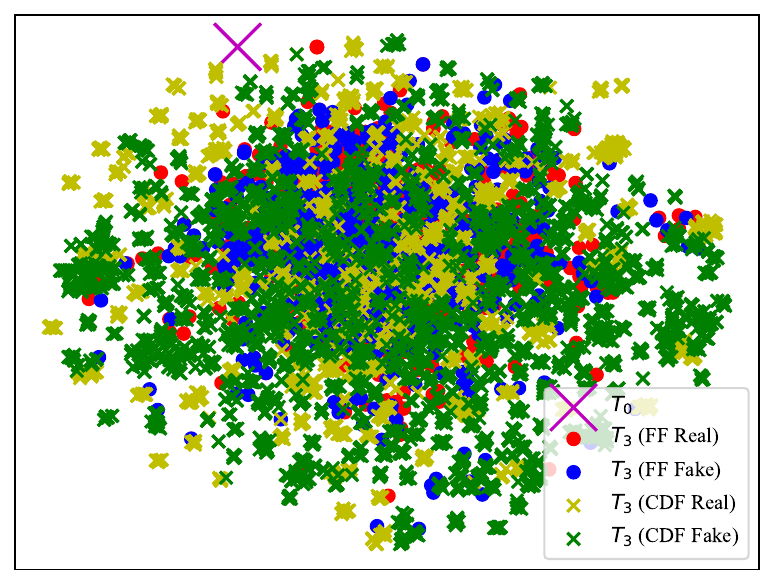}
	\caption{{The t-SNE visualization of textual features in FF and CDF datasets.}}
	\label{fig:tsne_text}
\end{figure}

In this experiment, we provide the t-SNE visualization of textual feature distributions. 
In RepDFD, textual features are generated using text templates and facial embeddings. 
As illustrated in Fig. \ref{fig:tsne_face}, the distributions of facial embeddings significantly do not overlap between the FF and CDF datasets due to the differences across domains. 
However, in Fig. \ref{fig:tsne_text}, we observe that the distributions of textual features (corresponding to $T_3$) for FF and CDF are more closely aligned, forming a common mid-domain. 
Therefore, we speculate that the textual encoder $E_T$ can squeeze different domains into a common mid-domain, which can effectively enhance the learning of our visual prompt $\delta$ by reducing domain discrepancies.

\begin{table}
	\centering
	\begin{tabular}{c|ccccc}
		\toprule
		$p$ &\#Para &CDF &Wild &DFDCP &Avg  \\
		\midrule
		12 &0.031 M &85.28 &85.25 &89.82 &86.78 \\
		23 &0.055 M &85.55 &87.01 &90.88 &87.81 \\
		34 &0.078 M &88.41 &87.73 &90.68 &\textbf{88.94} \\
		45 &0.097 M &80.64 &82.68 &90.37 &84.56 \\
		56 &0.113 M &82.17 &85.99 &91.40 &86.49 \\
		67 &0.126 M &78.59 &80.05 &88.65 &82.43 \\
		78 &0.137 M &74.18 &70.10 &86.90 &77.06 \\
		\bottomrule
	\end{tabular}
	\caption{The generalization performance involves different $p$ of the visual prompt. All models were trained on FF++ (DF).}
	\label{tab:width}
\end{table}



\end{document}